\documentclass{article}

    \PassOptionsToPackage{numbers, sort, compress}{natbib}

\usepackage[main, final]{neurips_2026}

\usepackage{hyperref}       
\usepackage{url}            
\usepackage{booktabs}       
\usepackage{amsfonts}       
\usepackage{nicefrac}       
\usepackage{microtype}      
\usepackage{xcolor}         
\usepackage{graphicx}       
\usepackage[table]{xcolor}
\usepackage{bm}
\usepackage{multirow}
\usepackage{comment}
\usepackage{etoc}
\usepackage{bibunits}
\defaultbibliographystyle{plainnat}

\usepackage[utf8]{inputenc} 
\usepackage{CJKutf8}
\usepackage[T1]{fontenc}    

\usepackage{pifont}
\newcommand{\cmark}{\textcolor{green!60!black}{\ding{51}}}
\newcommand{\xmark}{\textcolor{red!70!black}{\ding{55}}}

\title{Towards Real-Time Full-Waveform LiDAR Transformers via Intensity-Guided Token Reduction and Physics-Aware Augmentation}

\author{%
  Kotaro Oishi\thanks{These authors contributed equally and share first authorship. Either author may list their name first in CVs and other academic materials.} \\
  \texttt{koutarou00140805@keio.jp}\\
  Keio University \\
  \And
  Kazuma Ikeda\footnotemark[1] \\
  \texttt{kazu2080@keio.jp}\\
  Keio University \\
  \And
  Ryosei Hara \\
  \texttt{ryosei\_hara@keio.jp}\\
  Keio University \\
  \And
  Ryo Yoshida \\
  \texttt{ryo-yoshida@keio.jp}\\
  Keio University \\
  \And
  Mariko Isogawa\thanks{These authors are co-corresponding authors.} \\
  \texttt{mariko.isogawa@keio.jp}
  \\
  Keio University \\
  \And
  Kentaro Yoshioka\footnotemark[2] \\
  \texttt{kyoshioka47@keio.jp} \\
  Keio University \\
}

\begin{document}
\begin{CJK}{UTF8}{min}

\maketitle

\begin{abstract}
LiDAR is a critical sensor for autonomous driving and robotics, yet it remains vulnerable to adverse conditions such as fog, rain, transparent objects, and low-reflectance surfaces. Full-waveform LiDAR (FWL) addresses these limitations by capturing the complete return waveform as a histogram, preserving rich reflection characteristics including multi-path echoes and weak signals. While recent Transformer-based approaches have demonstrated promise on FWL data, two fundamental bottlenecks remain: high computational cost from redundant tokens in low-intensity background regions, and limited generalization due to scarce annotated training data.
We address both bottlenecks by exploiting the physical structure inherent to FWL data. First, we propose FWL-ToPM, a token pruning and merging mechanism that leverages intensity distributions to selectively reduce tokens while retaining critical waveform content. Second, we introduce FWLAug, a data augmentation framework designed as a regularization method that preserves waveform structure, retaining the Time-of-Flight waveform of each ray and the frustum geometry, 
to improve generalization to unseen scenes.
Evaluated on the Ghost-FWL benchmark, FWL-ToPM achieves over 8$\times$ speedup at real-time throughput alongside a nearly 9-point F1 gain over the prior state of the art, establishing the first real-time FWL Transformer that simultaneously improves accuracy. FWLAug further improves performance on both the standard benchmark and an out-of-distribution test set of unseen scenes. Together, these methods push FWL-based perception toward practical deployment.
The code and the OOD dataset are publicly available and can be accessed via the project page: \url{https://keio-csg.github.io/FWL-ToPM}

\end{abstract}

\section{Introduction}
\label{sec:intro}
\vspace{-2mm}
LiDAR is a key sensor for autonomous driving, robotics, and mapping, measuring 3D geometry via Time-of-Flight (ToF) of reflected laser pulses. Conventional LiDAR, however, records only the peak intensity of each pulse, and therefore cannot capture complex reflection characteristics such as multi-path returns from glass or backscatter from fog and rain. These limitations cause spurious point generation (ghosts and noise), degrading object detection and SLAM performance and introducing safety risks.

Full-waveform LiDAR (FWL) addresses these limitations by recording the complete temporal intensity waveform of each pulse as a histogram, thereby preserving rich reflection characteristics including multi-path echoes and weak-signal components. Transformer-based FWL models (FWLTs) exploit this richer representation to enable tasks fundamentally beyond conventional LiDAR, including ghost detection~\cite{ikeda2026ghostfwl} and fog-robust point cloud reconstruction~\cite{scheuble2025lidar}. Despite this promise, two fundamental challenges remain unresolved for practical FWLT deployment.

The first challenge is real-time inference. Because FWL records a complete histogram for every pixel, its data volume is far larger than that of a point cloud, yet the vast majority consists of low-intensity background regions carrying no object information. This redundancy severely inflates computational cost, limiting existing FWLTs to inference rates well below real-time requirements even on high-end GPUs~\cite{ikeda2026ghostfwl}. Existing token reduction methods~\cite{bolya2023token,rao2021dynamicvit}, designed for camera images and videos, assume informative content is broadly distributed across 2D space and thus indiscriminately discard the sparse high-intensity tokens that encode object-relevant reflections in FWL data. Exploiting FWL's physical structure for token reduction requires a dedicated mechanism.

The second challenge is generalization. FWL observations vary substantially across acquisition environments, object configurations, and reflectance conditions, yet publicly available data remains scarce. Minority classes such as glass and water lack sufficient samples to learn their characteristic reflection signatures, pushing models to overfit dominant classes. Standard image and point cloud augmentations ignore FWL's ToF waveform structure and frustum geometry and thus distort the waveform of each ray. A dedicated augmentation framework that respects these constraints is essential.

We address both challenges with two methods designed around the physical structure of FWL data. FWL-ToPM reduces computational cost by pruning redundant background tokens based on intensity distributions and merging the remainder, while preserving critical waveform content. FWLAug expands training diversity while respecting FWL's ToF waveform structure and frustum geometry. Together, these methods advance FWLTs toward practical deployment by simultaneously achieving real-time operation and improved generalization to unseen environments.
Our contributions are as follows:
\begin{itemize}
    \item We propose \textbf{FWL-ToPM}, a token pruning and merging mechanism tailored to FWL data. By leveraging intensity distributions to prune low-intensity background tokens and merge the remainder, FWL-ToPM reduces computational cost while retaining critical waveform content. It achieves an 8$\times$ speedup (53.2 FPS) and an approximately 9-point F1 improvement (0.503$\rightarrow$0.592), establishing the first real-time FWLT.

    \item We find that, unlike image-based Transformers where aggressive pruning destabilizes training~\cite{rao2021dynamicvit,liang2022evit}, even pruning up to 70\% of tokens can consistently improve FWLT accuracy. Removing low-intensity background tokens allows the model to focus on object-relevant signals that are otherwise diluted by uninformative waveform regions.

    \item We propose \textbf{FWLAug}, a data augmentation framework tailored to FWL data. FWLAug expands training diversity while preserving the ToF waveform structure and frustum geometry that standard image and point cloud augmentations violate.

    \item We construct an out-of-distribution (OOD) test set captured entirely outside the Ghost-FWL~\cite{ikeda2026ghostfwl} recording locations, on which FWLAug achieves the best F1-mean among all compared configurations.
    
\end{itemize}

\section{Related Work}
\label{sec:related}
\vspace{-2mm}

\paragraph{Deep Learning Methods for Full-waveform LiDAR.}
Conventional LiDAR internally selects several high-intensity peak positions for each laser pulse and stores data in a point cloud format ($x \times y \times z$), recording only the 3D coordinates of detected points.
In contrast, Full-waveform LiDAR (FWL) captures complete temporal light intensity information at each pixel of the receiver, which is stored in the FWL data format ($u \times v \times t$), where u and v denote the azimuth and elevation of each ray and t denotes the ToF bin.
From the complete histogram at each pixel, arbitrary peaks can be selected and converted into 3D coordinates, enabling the use of all waveform information and providing richer representations.
In recent years, various FWL-based studies have attracted attention, taking advantage of its characteristics, such as Classification~\cite{chehata2009airborne, liao2018tree, zorzi2019conv, shinohara2019fwnetae, shinohara2020fwnet, shinohara2020semantic}, 3D Object Detection~\cite{deziel2021pixset}, Jamming Recovery~\cite{yoshida2026neuralreconstructionlidarpoint}, Fog Removal~\cite{scheuble2025lidar, wallace2020adverse}, Ghost Removal~\cite{ikeda2026ghostfwl}.

While early FWL-based methods mainly used CNN-based networks, recent advancements in Transformers in the vision field~\cite{dosovitskiy2021vit, han2023visualtransformer, papa2024efficientvit} have led to the adoption of Transformer-based networks~\cite{scheuble2025lidar, ikeda2026ghostfwl}.
Scheuble et al.~\cite{scheuble2025lidar} proposed a Neural Digital Signal Processor (Neural DSP) using U-Net spatio-temporal transformers with FWL data as input, enabling robust point cloud reconstruction in low SNR environments such as fog.
However, the Neural DSP partitions tokens only along the time axis while treating spatial dimensions as $1{\times}1$, causing computational cost to scale poorly with spatial resolution and limiting throughput to 2 FPS.
Ikeda et al.~\cite{ikeda2026ghostfwl} proposed Ghost-FWL, a large-scale annotated FWL dataset for ghost detection and FWL-MAE, a self-supervised pre-training method for FWL data.
FWL-MAE treats the time axis as a single tube and applies a $16 \times 16$ patch division in the spatial direction to reduce computational complexity. Despite this, FWL-MAE reaches only 6 FPS, falling well short of 
the 10 Hz required for real-time operation.

Computational cost thus remains a fundamental bottleneck for FWLTs. We address this by introducing intensity-guided token pruning and merging tailored to FWL's sparse structure, targeting the ghost detection task in Ghost-FWL~\cite{ikeda2026ghostfwl} as a representative real-time perception challenge.

\vspace{-1mm}
\paragraph{Token Pruning and Merging.}
LiDAR-based systems require real-time performance, whereas deep learning–based methods tend to have high computational costs and low throughput. 
There is a trade-off between the accuracy and inference time of deep learning models. 
In particular, for Transformers, efficient vision transformers~\cite{liu2021swin, cai2023efficientvit, papa2024efficientvit} and token reduction methods such as pruning~\cite{rao2021dynamicvit, kong2022spvit, haurum2023which, liu2024revisiting, li2025mutual} and merging~\cite{bolya2023token,bolya2023tomesd,lu2025toma, li2024vidtome, lu2023contentaware, norouzi2024algm, pollard2025video, tran2025howmany} have been widely studied to reduce computational cost while maintaining accuracy.
Recently, methods that combine pruning and merging have also been proposed~\cite{kim2024token, mano2025pruneandmerge, bonnaerens2023learned}. However, most of these studies focus on token reduction for images or videos, and their application to FWL data remains unexplored.
These methods assume that significant information is widely distributed in 2D space, whereas FWL data has a sparse structure in 3D space where peaks appear at only a few locations and the rest are almost zero, making them unsuitable for FWL data.

In FWL data, regions around peaks contain more important information, while the majority of the data consists of intensities close to zero, making regions outside the peaks redundant and increasing computational cost.
Therefore, we propose FWL-ToPM, a token pruning and merging method tailored to FWL data that enables faster inference while maintaining high classification performance. 

\vspace{-1mm}
\paragraph{Augmentation for Full-waveform LiDAR.}

Augmentation is widely used to improve generalization by introducing diversity into limited training data, with extensive work spanning image recognition and 3D point clouds. For images, geometric and intensity transforms such as CutMix~\cite{yun2019cutmix} have been proposed, while 3D point cloud methods account for the geometric structure of the data. Mix3D~\cite{nekrasov2021mix3d} generates new training samples by blending multiple 3D scenes, and PolarMix~\cite{xiao2022polarmix} exploits the polar coordinate structure of LiDAR scans for rotation and region swapping.

No augmentation method has been designed specifically for FWL data. 
FWL data retains a continuous temporal waveform along each ray direction and forms a frustum-shaped observation structure with the LiDAR origin as its apex. Applying existing image or point cloud augmentations directly disrupts the ToF waveform structure and frustum geometry, causing unnatural distortions in object shape and distance. Augmentations proposed for ToF-based transient imaging~\cite{su2018deep,isogawa2020optical} are limited to basic operations such as noise addition and rotation, which are insufficient to improve generalization for FWL data.

\section{Method}
\label{sec:method}
\vspace{-2mm}
We propose a fast and robust framework for estimating the class logits $\bm{y} \in \mathbb{R}^{H \times W \times T \times C}$ of each voxel from FWL data $\bm{V} \in \mathbb{R}^{H \times W \times T}$, even for scenes that are not included in the training data.
Here, $H$, $W$, and $T$ denote the spatial and temporal resolutions, respectively; $H$ and $W$ correspond to the elevation ($v$) and azimuth ($u$) axes of the ($u \times v \times t$) format, and $T$ to the ToF ($t$) axis, and $C$ denotes the number of target classes, such as \textit{Noise}, \textit{Object}, \textit{Glass}, and \textit{Ghost}.

The proposed framework consists of FWL-ToPM, which enables fast and accurate FWL recognition, and FWLAug, which improves generalization performance for unseen scenes.
FWL-ToPM focuses on the spatial sparsity of signal intensity specific to FWL data and achieves efficient feature extraction by reducing redundant tokens.
FWLAug generates diverse training samples of waveform structures, object arrangements, and minority classes while preserving the ToF-direction waveform structure specific to FWL data and the frustum geometry based on the LiDAR viewpoint.
This increases the diversity of the training data and enables the model to learn more robust representations for unseen scenes.
The details of FWL-ToPM and FWLAug are described in \S\ref{ssec:topm} and \S\ref{ssec:augmentation}, respectively.

\subsection{FWL-ToPM}
\label{ssec:topm}
\begin{figure}[tp]
  \begin{center}
    \includegraphics[width=1.0\linewidth]{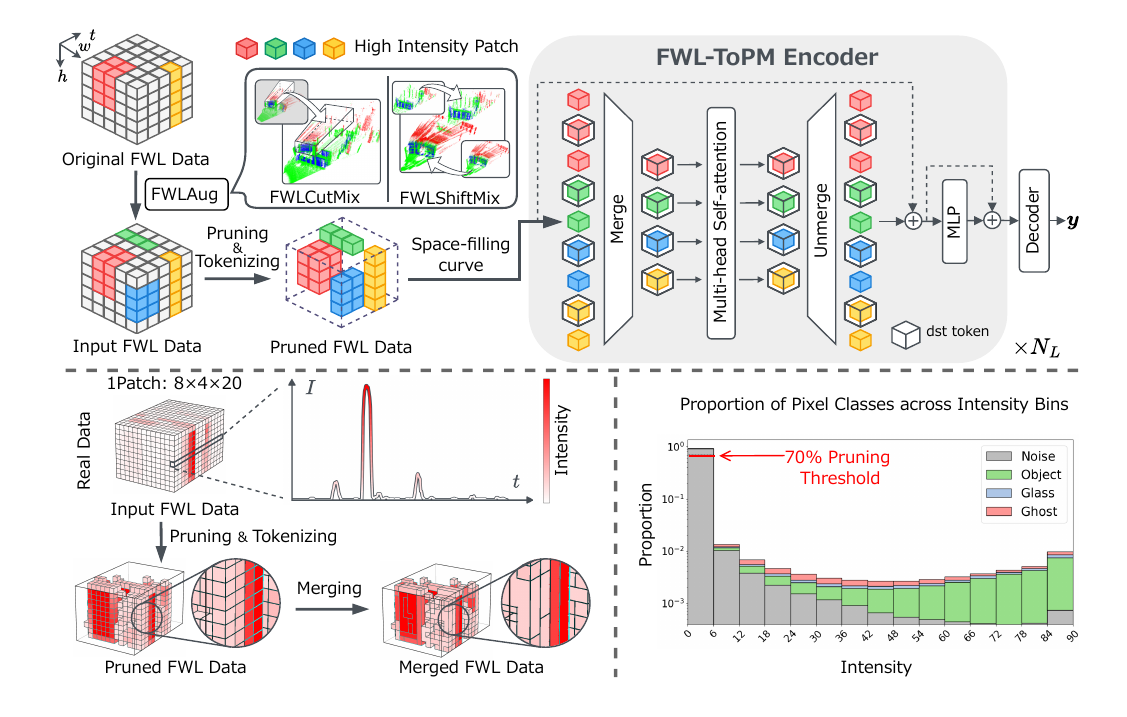}
    \caption{Overview of the proposed framework. \textbf{(Top)} During training, FWLAug augments FWL samples while preserving ToF waveform structure and frustum geometry. High-intensity patches are selected by pruning, spatially reordered along a space-filling curve, and processed by the \textbf{FWL-ToPM} Encoder, which merges redundant tokens before self-attention and unmerges afterward across $L$ Transformer blocks.  
    \textbf{(Bottom Left)} Pruning removes low-intensity background tokens, and merging aggregates the remainder, concentrating computation on object-relevant waveform content.
    \textbf{(Bottom Right)} Proportion of pixel classes across intensity bins in FWL data. Most regions in FWL data correspond to noisy background, and our pruning strategy removes these dominant low-intensity regions.}
    \label{fig:dataset_scene}
  \end{center}
  \vspace{-6mm}
\end{figure}
Recently, the effectiveness of Transformer architectures has been reported for the analysis of FWL data~\cite{scheuble2025lidar, ikeda2026ghostfwl}.
However, existing methods are largely based on Vision Transformers (ViTs), which were originally designed for images, and therefore do not sufficiently account for the characteristics of FWL data.
Unlike images, FWL data are characterized by spatially sparse peaks with high signal intensity, while most regions are occupied by low-intensity signals that contribute little to the classification task.
As a result, applying uniform ViT-based tokenization generates many uninformative tokens, which can obscure important high-intensity regions.
Consequently, a large portion of computational resources is spent on ineffective tokens.
Moreover, since low-intensity tokens dominate the representation, an imbalance in feature representations occurs, leading to degraded recognition performance.
These issues cannot be resolved simply by applying token merging methods for 2D images~\cite{bolya2023token, bolya2023tomesd}, which are designed under the assumption that information is relatively homogeneous across pixels.

To address these issues, we propose FWL-ToPM, which explicitly exploits the signal characteristics of FWL data to optimize token representations specifically for FWL Transformers.
Specifically, FWL-ToPM combines
(1) signal-intensity-based token pruning and
(2) spatial-structure-aware token merging
to achieve both computational efficiency and recognition performance.

\vspace{-2mm}
\paragraph{Model Architecture.}
FWL-ToPM uses a Vision Transformer (ViT)~\cite{dosovitskiy2021vit} as the backbone for voxel-wise classification.
However, the main contribution of our method lies in its token selection and aggregation mechanisms, which are applicable to any token-based models.
First, the input frame $\bm{V} \in \mathbb{R}^{H \times W \times T}$ is divided into 3D patches of patch size $\bm{p} = (p_H, p_W, p_T)$.
Then, each patch is projected into an $E$-dimensional patch token by a 3D convolution with both kernel size and stride set to $\bm{p}$.
A 3D positional embedding corresponding to the position of each patch is added to these tokens, yielding the token sequence used as the input to the encoder.
Next, these tokens are input into a Transformer encoder composed of $N_L$ blocks.
Each Transformer block consists of a multi-head self-attention layer with $N_h$ attention heads and an MLP.
When pruning and merging are not introduced, the encoder processes all patch tokens as they are, and $E$-dimensional features are extracted for each patch at the final layer.
Finally, a single-layer linear decoder is applied to each token to predict class logits for the voxels within the corresponding patch.
These logits are rearranged according to the original patch layout to obtain the final output $\bm{y} \in \mathbb{R}^{H \times W \times T \times C}$.
The following describes the details of the proposed token pruning and token merging.

\paragraph{Intensity-Based Pruning.}

In FWL data, peaks in the LiDAR histogram with high signal intensity are sparsely distributed, while most low signal intensity regions correspond to noise that does not contribute to the classification task.
Therefore, our method evaluates the importance of each patch based on its maximum signal intensity and performs token pruning to remove low signal intensity patches in advance.
Specifically, the procedure is as follows: 
First, we store the maximum signal intensity within the $i$-th patch as $m_i$.
Next, all patches are sorted in ascending order according to $m_i$, and the bottom $\rho_{\mathrm{low}}$ proportion of patches are assigned to ${\bm{P}_{\mathrm{low}}}$ as removal targets, while the remaining patches are assigned to ${\bm{P}_{\mathrm{high}}}$.
Only the patches belonging to ${\bm{P}_{\mathrm{high}}}$ are tokenized and fed into the encoder.
Since the removed patches do not pass through the encoder, the decoder cannot predict class logits for the corresponding patch regions.
Therefore, during training, the loss is computed by excluding the losses for the pruned patch regions, while during inference, all voxels in those patch regions are predicted as the noise label.

Our proposed pruning method consists only of patch selection and modification of the token pathway.
It does not modify the weights of the patch embedding, encoder, and decoder.
Therefore, this pruning method can be applied to a pretrained model at inference time without additional training.

\vspace{-2mm}
\paragraph{Sparse-Aware Token Merging.}
Although pruning can remove most highly redundant tokens, redundant tokens may still remain among the patches in ${\bm{P}_{\mathrm{high}}}$ after pruning.
Therefore, we apply token merging to the tokens in ${\bm{P}_{\mathrm{high}}}$ to further reduce the number of tokens used in attention computation.
Token merging is a method that selects representative destination tokens (dst tokens) from a token sequence and aggregates the remaining source tokens (src tokens) into the dst tokens based on feature similarity.
To make token merging work effectively, it is necessary to appropriately sample dst tokens in a way that reflects the characteristics of the data.

In the case of FWL data, spatially neighboring patches are likely to contain similar waveform information originating from the same object or reflection region.
Therefore, the dst tokens used for merging should be distributed across the original 3D space rather than being biased toward specific local regions.
A key challenge in applying token merging after pruning is that $\bm{P}_{\mathrm{high}}$ forms a \textit{sparse and irregular} subset of the original 3D patch grid. Existing dst token selection methods divide the patch grid into fixed-size regions and sample one dst token per region~\cite{bolya2023tomesd,zhang2024ninetimes,sun2025asymrnr,zhang2025trainingfree}; these methods assume a dense, uniform token distribution and fail when applied to a post-pruning sparse set, as regions may contain no tokens at all.

We resolve this by serializing ${\bm{P}_{\mathrm{high}}}$ along a space-filling curve prior to dst selection.
A space-filling curve maps the 3D patch grid to a 1D sequence such that patches that are close in 3D space are placed near each other in the sequence.
Using this property, we compute the position along the curve of each patch token in ${\bm{P}_{\mathrm{high}}}$ from its coordinates in the 3D patch grid before pruning, and reorder the tokens accordingly. 
This constructs a token sequence that preserves spatial relationships even when gaps are introduced into the token sequence by pruning.
We then select dst tokens at equal intervals from the serialized token sequence according to the dst token ratio $\rho_d$, and treat the remaining tokens as src tokens.
Next, we compute the cosine similarity between the features of each src token and each dst token, and assign each src token to the dst token with the highest similarity.
Then, based on the merge ratio $\rho_m$, src tokens with high similarity are merged by averaging them with their corresponding dst tokens.

This serialization does not depend on a particular curve, as long as the curve preserves spatial proximity in the 1D sequence. In our experiments, we adopt the Morton order~\cite{morton1966computer}, which is obtained by interleaving the coordinate bits of the three axes for each patch. Serializing 3D data with space-filling curves has also been adopted in Point Transformer V3 (PTv3)~\cite{PTv3}, which orders point clouds along Z-order and Hilbert curves to partition them into patches for local attention. Our approach shares the same principle of preserving spatial proximity in a 1D sequence, but differs in its input and purpose: we serialize the sparse and irregular set of 3D waveform patches that remain after intensity-based pruning, in order to select spatially distributed dst tokens for merging.

Since token merging is performed to reduce the computational cost of attention, after the attention computation, the number of tokens must be restored by reassigning the merged destination features to each token, thereby maintaining consistency with the residual connections.
Therefore, after the attention computation, for each merged src token, we copy the feature of the dst token into which it was merged and redistribute it back to the original token position.
Our merging method also does not change the parameter weights, and can be applied to pre-trained models without additional training.

\subsection{FWLAug}
\label{ssec:augmentation}

\begin{figure}[tp]
  \begin{center}
    \includegraphics[width=0.9\linewidth]{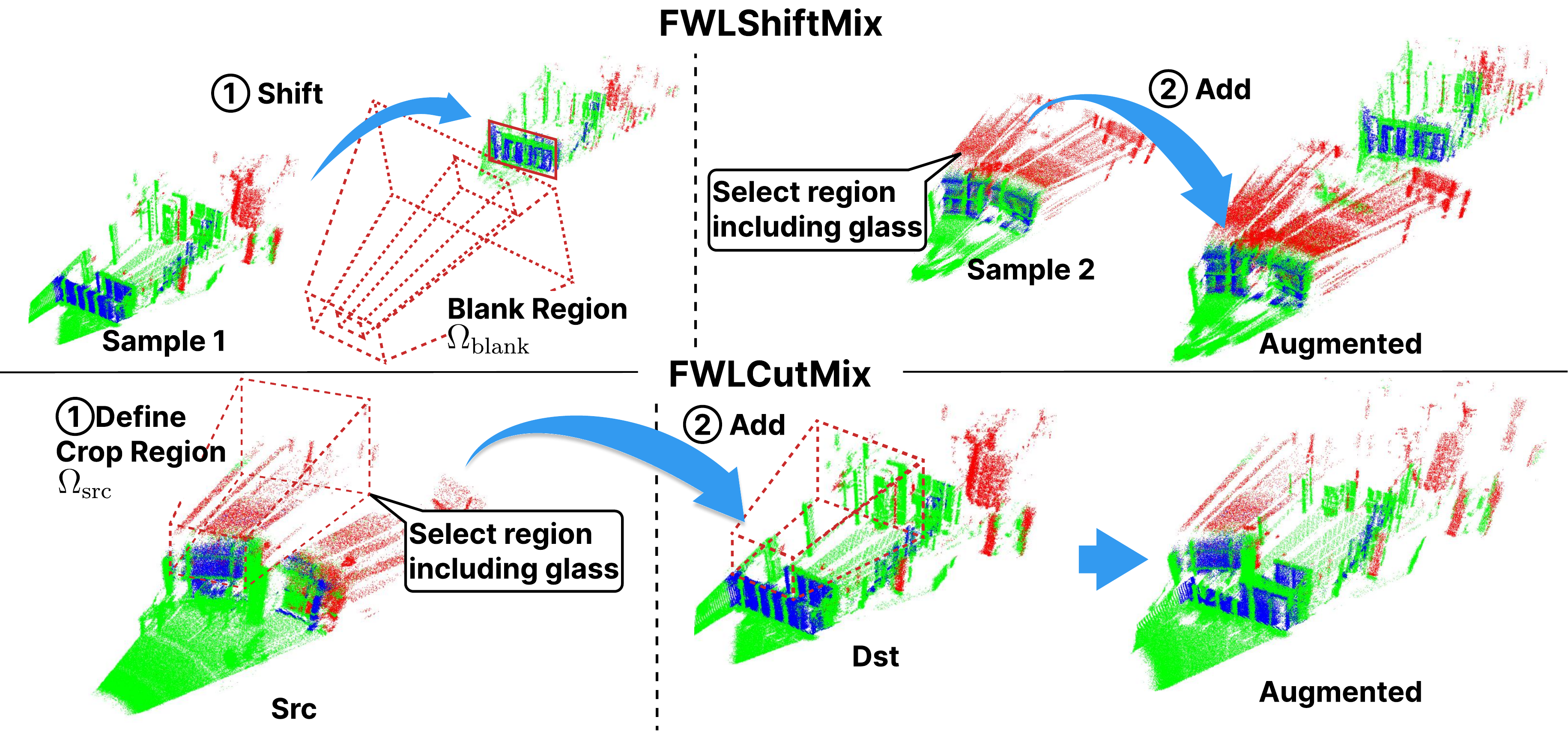}
    \caption{Illustration of FWLAug. \textbf{FWLShiftMix} (top): an object in Sample~1 is shifted away from the LiDAR, creating a blank region $\Omega_{\mathrm{blank}}$. A glass-containing region $\Omega_{\mathrm{src}}$ is then selected from Sample~2 and pasted into the blank, producing an augmented scene in which the waveform structure of each ray is preserved. \textbf{FWLCutMix} (bottom): a crop region $\Omega_{\mathrm{src}}$ containing minority class (glass) is selected from a source frame and pasted into a destination frame at a random location. Both methods prioritize glass-containing regions to mitigate class imbalance.}
    \label{fig:augmentation}
  \end{center}
  \vspace{-6mm}
\end{figure}

We propose FWLAug, an augmentation framework designed around the physical structure of FWL data (Figure~\ref{fig:augmentation}). In FWL data $\bm{V} \in \mathbb{R}^{H \times W \times T}$, each per-pixel histogram $\bm{Q}_{h,w}$ encodes the temporal intensity response of a laser pulse, preserving time-of-flight delays, reflectance, and multi-path waveform characteristics. 
Since each $\bm{Q}_{h,w}$ corresponds to a ray from the LiDAR origin, the full FWL frame forms a frustum-shaped observation structure. These properties impose two requirements on any FWL augmentation: (1) the $T$-axis waveform structure of each $\bm{Q}_{h,w}$ must be preserved, and (2) consistency with frustum geometry must be maintained. 
While FWLAug is not intended to serve as a strict optical simulator that reproduces the complete light transport of the augmented scene, it is strategically designed as an effective regularization method. By intentionally focusing on preserving the ray-wise waveform structure, it efficiently increases training diversity and the representation of minority classes while satisfying the two requirements above.

FWLAug comprises two augmentations that address independent challenges. 
FWLCutMix alleviates minority-class under-representation by prioritizing glass-containing regions during patch pasting, while FWLShiftMix expands diversity in observation distance and position by translating objects in 3D space. Used together, they improve generalization to unseen scenes while preserving in-domain accuracy.

\vspace{-3mm}
\paragraph{FWLShiftMix.}
Shifting objects away from the LiDAR increases distance diversity but creates blank regions due to the compression effect. Zero-filling these regions introduces unrealistic data distributions absent from real FWL scenes, causing the model to learn spurious artifacts. FWLShiftMix instead fills the blank regions with FWL data from another scene, producing synthetic samples that preserve the waveform structure of each ray.

Concretely, FWLShiftMix proceeds in three steps (Figure~\ref{fig:augmentation}).
First, each pixel's peak position $t^{*}_{h,w} = \arg\max_t \bm{Q}_{h,w}(t)$ is converted to a 3D point $\bm{r}_{h,w}$ via ToF geometry. Second, we translate each point along its LiDAR ray direction. Let $\bm{u}_{h,w} = \bm{r}_{h,w} / \|\bm{r}_{h,w}\|_2$ denote the unit ray direction and let $\Delta d \sim \mathcal{U}(-3, 15)$ denote the sampled radial displacement in meters.
The translated point is computed as $\bm{r}'_{h,w} = \bm{r}_{h,w} + \Delta d \bm{u}_{h,w}$ and is then reprojected back into
the range view. The waveform at each reprojected pixel is shifted along the $T$ axis according to the corresponding range change, so that the histogram remains consistent with the new object position.
Unlike a direct tensor shift, this 3D-space manipulation preserves object geometry.
Third, the blank region $\Omega_{\mathrm{blank}}$ left by the shifted object is filled
with waveform data from another scene $\bm{V}^{\mathrm{src}}$. We select a
glass-containing source region $\Omega_{\mathrm{src}}$ from $\bm{V}^{\mathrm{src}}$
and define a mapping $\phi : \Omega_{\mathrm{blank}} \to \Omega_{\mathrm{src}}$
that assigns each blank pixel to a source pixel. For each pixel in
$\Omega_{\mathrm{blank}}$, we copy the entire temporal waveform from its mapped
source pixel in $\Omega_{\mathrm{src}}$. This fills the unrealistic blank region while
preserving the ToF waveform structure of the pasted FWL data. Prioritizing glass-containing regions further increases minority-class frequency during training.

\vspace{-2mm}
\paragraph{FWLCutMix.}
FWLCutMix extends CutMix~\cite{yun2019cutmix} to FWL data by combining a destination sample $\bm{V}_{\mathrm{dst}} \in \mathbb{R}^{H \times W \times T}$ and a source sample $\bm{V}_{\mathrm{src}} \in \mathbb{R}^{H \times W \times T}$ to generate a synthetic scene. The crop size is determined by a mixing ratio $\lambda$ sampled randomly: $ell_H = \lfloor H\sqrt{1-\lambda} \rfloor$, $ell_W = \lfloor W\sqrt{1-\lambda} \rfloor$.
A minority-class region $\Omega_{\mathrm{src}}$ is preferentially selected from $\bm{V}_{\mathrm{src}}$ and pasted at a random location in $\bm{V}_{\mathrm{dst}}$ to mitigate class imbalance. The waveform $\bm{Q}_{h,w}$ is pasted in its entirety along the $T$ axis, preserving the temporal waveform structure.

\section{Experimental Settings}
\label{sec:experiment}
\vspace{-2mm}


\paragraph{Implementation details.}
All methods are trained using the Ghost-FWL dataset~\cite{ikeda2026ghostfwl}, the only 
publicly available FWL dataset with peak-level annotations to date.
The raw FWL data in Ghost-FWL has a size of $(H \times W \times T) = (512 \times 400 \times 700)$.
First, the top and bottom 88 bins corresponding to reflections from the ceiling and floor and the front 25 bins containing noise from internal sensor reflections were removed.
We also removed the rear 375 bins, since the data contains only a few object peaks in the rear region, resulting in a full-frame size of $(H \times W \times T) = (336 \times 400 \times 300)$.
Furthermore, due to GPU memory limitations in the comparison methods, we use a crop corresponding to one quarter of the full frame, with size $(H \times W \times T) = (168 \times 200 \times 300)$, as the input for all methods in order to keep the input size consistent.
For the Vision Transformer, the patch size is set to $(p_H \times p_W \times p_T) = (8 \times 4 \times 20)$, the number of Transformer blocks $N_L$ is set to 6, and the number of attention heads $N_h$ is set to 6.
The feature dimension $E$ is set to 240, and the ratio of dst tokens $\rho_d$ is set to 0.1.
Following Ghost-FWL~\cite{ikeda2026ghostfwl}, we use the focal loss~\cite{lin2017focal} as the training objective.
We use AdamW~\cite{Loshchilov2017DecoupledWD} as the optimizer, set the learning rate to $1.0 \times 10^{-4}$, and train the model for 50 epochs.
All training and evaluation were performed on a computer equipped with an Intel Core i9-14900K CPU and a single RTX 6000 Ada Generation GPU.
Further details are provided in the supplemental material.

\vspace{-1mm}
\paragraph{Metrics.}
Following several prior studies~\cite{shinohara2020fwnet, shinohara2020semantic}, we evaluate performance using the mean F1 score over all classes excluding Noise (F1-mean).
To verify the inference speedup achieved by token pruning and merging, we report the number of full-size frames that can be inferred per second (FPS).

\vspace{-1mm}
\paragraph{Baseline Methods.}
To investigate the effectiveness of the proposed method, we compare it with three FWL-based methods: 
(1) the proposed method (Ours), 
(2) a Transformer-based model using pretraining, which is the state-of-the-art model for FWL-based ghost detection (Ikeda et al.)~\cite{ikeda2026ghostfwl}, and 
(3) a model obtained by attaching a classification head for ghost detection to the backbone of a spatio-temporal Transformer model with a U-Net architecture for digital signal processing using FWL (Scheuble et al.)~\cite{scheuble2025lidar}.
We also compare our method with pruning and merging methods used for image-based models: 
(1) DynamicViT~\cite{rao2021dynamicvit}, which learns whether each token should be pruned using an MLP and then performs pruning, and 
(2) ToMe~\cite{bolya2023token}, which divides tokens into two groups and merges them based on cosine similarity between tokens.

\vspace{-1mm}
\paragraph{Augmentation.}
At each training iteration, either FWLCutMix or FWLShiftMix is selected with equal probability (0.5). For FWLCutMix, 1--3 glass regions are sampled from a randomly selected training frame and pasted into the current frame. For FWLShiftMix, the shift magnitude is sampled uniformly from $[-3, 15]$ meters.

\noindent\textbf{OOD Test Set.}
To evaluate generalization, we construct an OOD test set at a location entirely separate from Ghost-FWL~\cite{ikeda2026ghostfwl} (see supplementary). The data were collected with the same FWL sensor platform and acquisition configuration as Ghost-FWL and annotated following the same protocol, so that the evaluation isolates changes in scene distribution rather than changes in sensor hardware or annotation procedure. The 500-frame set contains over 300 glass instances and scenes with multiple spatially separated glass regions not present in Ghost-FWL, providing a challenging benchmark for FWLT out-of-distribution evaluation.

\begin{table*}[t]
\centering
\caption{
Comparison with FWL-based methods and ablation of token pruning and merging components.
For FWL-ToPM, we set $\rho_{\mathrm{low}}=0.7$ and $\rho_m=0.9$.
FWL-ToPM achieves the highest F1 and improves throughput by 8$\times$ over the prior state of the art.
Our pruning and merging components each outperform their image-based counterparts (DynamicViT and ToMe) and their combination achieves the best F1--FPS trade-off.
All methods in this table are trained without FWLAug.
}
\vspace{2mm}
\label{tab:fwl_topm_results}
\begin{tabular}{lcccc}
\hline
Method & Pruning & Merging & F1-mean & FPS \\
\hline
Scheuble et al.~\cite{scheuble2025lidar} & -- & -- & 0.497 & 1.97 \\
Ikeda et al.~\cite{ikeda2026ghostfwl} & -- & -- & 0.503 & 6.54 \\
\hline
\hline
Ours w/o FWL-ToPM & -- & -- & 0.555 & 2.01 \\
\hline
DynamicViT~\cite{rao2021dynamicvit} & \cmark & -- & 0.480 & 20.0 \\
Ours w/Pruning & \cmark & -- & 0.591 & 20.3 \\
\hline
ToMe~\cite{bolya2023token} & -- & \cmark & 0.543 & 5.24 \\
Ours w/Merging & -- & \cmark & 0.565 & 18.0 \\
\hline
\hline
Ours w/FWL-ToPM & \cmark & \cmark & \textbf{0.592} & \textbf{53.2} \\
\hline
\end{tabular}
\end{table*}

\begin{figure}[t]
    \centering

    \includegraphics[width=\columnwidth]{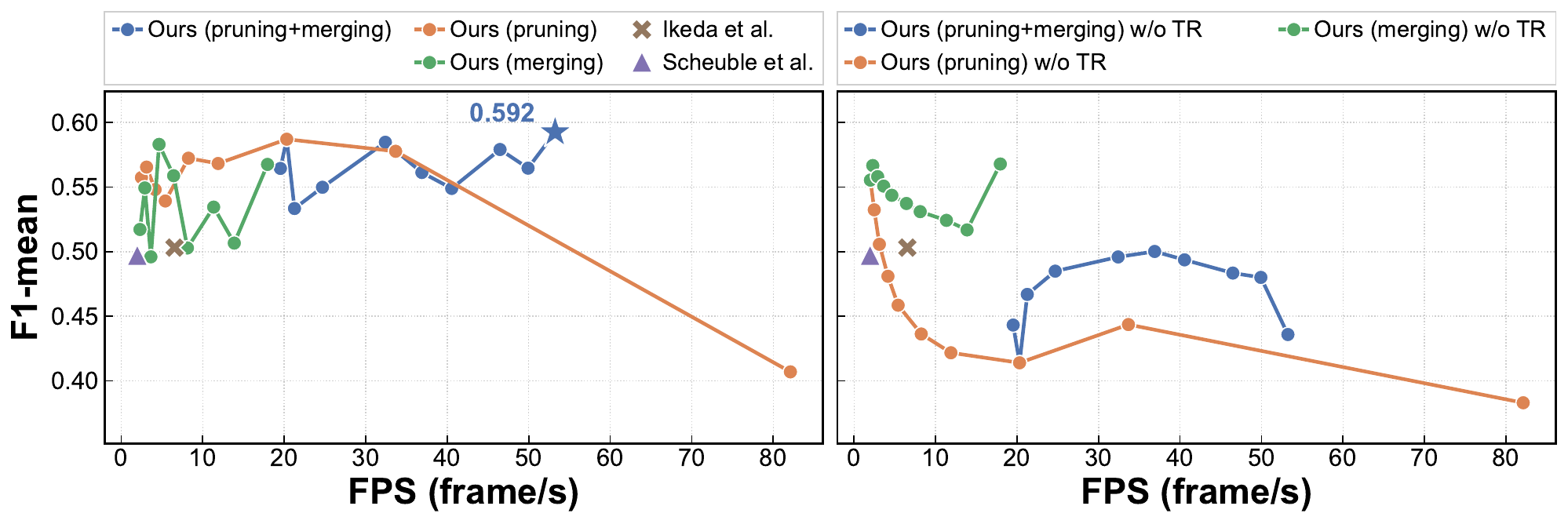}
    \vspace{-4mm}
    \caption{
    \textbf{(Left)} F1-mean vs. FPS trade-off for all compared methods.
    \textbf{(Right)} Comparison of F1-mean and FPS without additional training.
    For pruning and merging, $\rho_{\mathrm{low}}$ and $\rho_m$ are varied from 0 to 0.9, respectively.
    For pruning+merging, $\rho_{\mathrm{low}}$ is fixed to 0.7 while $\rho_m$ is varied.
    All models are trained without FWLAug.}
    \label{fig:combine}
    \vspace{-2mm}
\end{figure}

\begin{figure}[t]
    \centering

    \includegraphics[width=\columnwidth]{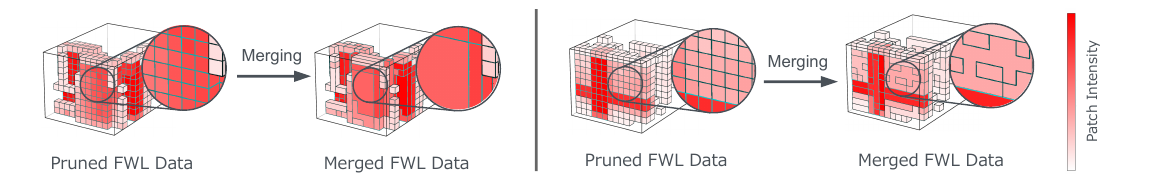}
    \vspace{-4mm}
    \caption{
    Visualization of token merging on two real Ghost-FWL scenes. Color indicates patch intensity (red: high, white: low). After pruning low-intensity background tokens, the remaining high-intensity tokens are merged by aggregating similar-feature source tokens into destination tokens, reducing the total token count while preserving object-relevant content. The zoomed-in regions show that merged tokens retain the spatial structure of reflected objects across diverse scenes.}
    \label{fig:vis}
    \vspace{-2mm}
\end{figure}

\section{Experiments and Results}
\label{sec:results}
\vspace{-2mm}
In this section, we conducted the following experiments.
To evaluate the effectiveness of the FWL-ToPM architecture, we conducted 
(1) a comparison with existing FWL-based recognition methods and (2) a comparison with image-based token reduction baselines. 
We then ablate (3) the pruning and merging components and the (4) FWL augmentations. 
We also (5) analyze the training-free applicability of token reduction, and (6) discuss limitations.

\paragraph{Comparison with existing FWL-based recognition methods.}
To evaluate the effectiveness of the FWL-ToPM architecture, we compared its accuracy and computational cost with those of existing FWL-based recognition methods.
The results are presented in Table~\ref{tab:fwl_topm_results} and the left part of Figure~\ref{fig:combine}.
Our FWL-ToPM simultaneously achieved a high F1-mean score of 0.592 and an inference speed of 53.2 FPS.
Compared with the baselines of Scheuble et al.~\cite{scheuble2025lidar} and Ikeda et al.~\cite{ikeda2026ghostfwl}, this corresponds to an approximately 9-point improvement in the F1 score and a speedup of more than 8$\times$ in inference speed, demonstrating that FWL-ToPM achieves both high accuracy and high efficiency.
A qualitative evaluation of the classification results for the comparison with existing FWL-based methods is provided in the supplementary material.

\vspace{-2mm}
\paragraph{Comparison with existing pruning and merging methods.}
To verify the effectiveness of the proposed pruning and merging methods, we compared them with existing methods designed for image-based Transformers.
The results are presented in Table~\ref{tab:fwl_topm_results}.
As a result, both our pruning and merging methods individually outperformed the existing methods, confirming that each component is independently effective.

\vspace{-2mm}
\paragraph{Ablation study on pruning and merging.}
To clarify the contribution of each component in the proposed method, we conducted an ablation study on pruning and merging.
The results are presented in Table~\ref{tab:fwl_topm_results} and the left part of Figure~\ref{fig:combine}.
As shown in Table~\ref{tab:fwl_topm_results}, the F1-mean score was 0.555 when neither component was applied, whereas it increased to 0.591 when only token pruning was applied and to 0.565 when only token merging was applied.
These results confirm that both components contribute to improving accuracy.

Moreover, as shown in the left part of Figure~\ref{fig:combine}, token pruning alone maintains high accuracy up to around 20 FPS, with an F1-mean of approximately 0.59.
However, when the token pruning ratio is increased to achieve a speed of around 80 FPS, the accuracy drops substantially.
This is likely because token pruning also removes high-intensity information that is essential for classification.
On the other hand, token merging alone suppresses the degradation in accuracy, but its ability to reduce computational cost is limited, resulting in a speed of only around 18 FPS.
By combining pruning and merging, the proposed method removes unnecessary background regions through pruning while aggregating the remaining effective waveform information through merging.
This complementary design mitigates the limitations of each individual component and enables the method to simultaneously achieve high accuracy and real-time performance exceeding 50 FPS.

\vspace{-2mm}
\paragraph{Ablation Study of FWLAug.}
Table~\ref{tab:fwl_aug} shows the contribution of each FWLAug component to both in-distribution and OOD performance. The full FWLAug achieves the largest OOD F1 gain (4.6-point over baseline), particularly on the minority Glass class (8.4-point), confirming improved generalization to unseen scenes. FWLShiftMix contributes more to in-distribution accuracy, while the combination of both drives the OOD improvement, suggesting complementary effects.

\vspace{-2mm}
\paragraph{Applicability to pretrained models.}
To evaluate the behavior of token reduction without additional training, we analyzed the performance when pruning and merging were directly applied to pretrained models.
The results are presented in the right part of Figure~\ref{fig:combine}.
When token pruning was applied without additional training, the accuracy dropped sharply as the reduction ratio increased.
This is likely because the model, which was trained using all tokens, also relies on information from low-intensity patches.
In contrast, when token merging was applied without additional training, the F1-mean remained above 0.5 even when the inference speed was increased to around 20 FPS, indicating that severe accuracy degradation did not occur.
This result indicates that the proposed merging method can effectively reduce computational cost without substantially degrading feature representations, even without additional training.

\paragraph{Limitations.}
While our proposed framework demonstrates significant improvements, we acknowledge several limitations that provide directions for future work:
\begin{itemize}
    \item \textbf{Risk of removing weak signals:} FWL-ToPM prunes low-intensity regions by regarding them as background or noise, and therefore may incorrectly remove useful weak signals, such as distant objects, low-reflectance objects, or weak reflections under adverse weather conditions. Since pruned patches are treated as noise during inference, such erroneous removal can directly lead to false negatives. Thus, adaptively determining $\rho_{\mathrm{low}}$, $\rho_m$, and $\rho_d$ according to the scene and signal distribution remains an important direction for future work.

    \item \textbf{Single sensor platform:} Our evaluation relies on a single FWL sensor platform: generalization is evaluated only on unseen scenes acquired with the same sensor platform, acquisition configuration, and annotation protocol as Ghost-FWL, and cross-sensor evaluation has not yet been conducted. Broader validation across different sensors, resolutions, and weather conditions remains necessary.
    
    \item \textbf{Scope of physical consistency in FWLAug:} FWLAug is a regularization method that preserves waveform structure rather than an optical simulator. Specifically, FWLShiftMix does not model distance-dependent intensity attenuation or recompute the optical interactions between the pasted glass regions and the shifted objects, and FWLCutMix does not guarantee consistent reflection paths across the entire scene because of spatial discontinuities at the paste boundaries.
    
    \item \textbf{System-level efficiency evaluation:} Our current efficiency evaluation is limited to model-level throughput. Sensor-to-decision latency, including sensor input, preprocessing, and system integration, as well as deployment on actual embedded robotic platforms, are important directions for future work.
\end{itemize}

\begin{table}[t]
\centering
\renewcommand{\arraystretch}{1.15}
\centering
\caption{Ablation study of FWLAug. To evaluate out-of-distribution generalization, we construct an OOD test set captured at a location entirely separate from Ghost-FWL~\cite{ikeda2026ghostfwl}. The full FWLAug achieves the best F1 on the OOD test set, particularly for the minority Glass class, confirming improved generalization to unseen scenes.}
\label{tab:fwl_aug}
\small
\begin{tabular}{lccccccc}
\hline
\multirow{2}{*}{Method} & \multirow{2}{*}{FWLCutMix} & \multirow{2}{*}{FWLShiftMix}
  & \multicolumn{2}{c}{Ghost-FWL~\cite{ikeda2026ghostfwl}} 
  & \multicolumn{2}{c}{OOD Test} \\
 & & & F1-mean & Glass & F1-mean & Glass \\
\hline
Ours w/o Aug
  & \xmark & \xmark & 0.592 & 0.352 & 0.406 & 0.491 \\
\hline
\multirow{3}{*}{Ours w/ FWLAug}
  & \xmark & \cmark & \textbf{0.622} & \underline{0.376} & 0.425 & 0.494 \\
  & \cmark & \cmark & \underline{0.617} & \textbf{0.385} & \textbf{0.452} & \underline{0.575} \\
  & \cmark & \xmark & 0.586 & 0.315 & \underline{0.434} & \textbf{0.606} \\
\hline
\end{tabular}
\vspace{-2mm}
\end{table}

\vspace{-2mm}
\section{Conclusion}
\label{sec:conclusion}
\vspace{-2mm}
We presented FWL-ToPM and FWLAug, two complementary methods that address the core bottlenecks of real-time inference and limited generalization for Full-Waveform LiDAR Transformers. FWL-ToPM exploits the intensity sparsity inherent to FWL data to prune redundant background tokens and merge the remainder, achieving 53.2 FPS with an 8$\times$ speedup and a 9-point F1 gain over the prior state-of-the-art — establishing the first real-time FWLT. A notable finding is that pruning up to 70\% of tokens consistently improves accuracy, suggesting that low-intensity tokens act as noise that dilutes object-relevant signals. FWLAug improves generalization through FWLCutMix and FWLShiftMix, both preserving FWL's ToF waveform structure and frustum geometry, with gains confirmed on an OOD test set of unseen scenes. Together, these advances bring real-time FWL perception a step closer to deployment in autonomous driving and robotics.

\paragraph{Acknowledgements and disclosures.}
We gratefully acknowledge Takahiro Kado, Ibuki Fujioka, and Taro Beppu at Sony Semiconductor Solutions for their generous support and for providing the Sony LiDAR sensor used in this work. The Sony LiDAR system played an essential role in enabling this research.
This research was supported in part by JST Next-generation Edge AI Semiconductor
JPMJES2515, JST CREST JPMJCR21D2, JPMJCR23M4, JST PRESTO JPMJPR22PA, JST ASPIRE JPMJAP2515, JST FOREST JPMJFR242I, JSPS KAKENHI 24K02940, and 25H01159.

{
\small
\bibliographystyle{plainnat}
\bibliography{main}
}

\clearpage
\appendix


\end{CJK}
\end{document}


\refstepcounter{part}
\addcontentsline{toc}{part}{Supplementary Material}

\begin{center}
  {\LARGE \bfseries Supplementary Material}
\end{center}

\makeatletter
\setcounter{NAT@ctr}{0}
\makeatother

\begingroup
\hypersetup{linkcolor=black}
\etocsettocstyle{\section*{Contents}}{}
\localtableofcontents
\endgroup

\section{Implementation Details}
\label{sec:implementation}

\subsection{Training}
We evaluated our method on the Ghost-FWL dataset~\cite{ikeda2026ghostfwl}.
Following the original data split, we assigned Scenes 001, 003, 004, 005, 006, 008, and 010 to the training and validation set, and Scenes 002, 007, and 009 to the test set. 
The original dataset provided by Ikeda et al.~\cite{ikeda2026ghostfwl} contains 13,853 training frames and 2,994 validation frames.
To reduce computational cost, we randomly sampled one-third of the data from both sets while preserving the scene distribution.
Consequently, our final subsets consist of 4,617 training frames and 998 validation frames. 
For the test set, we did not perform any sampling and used all 1,427 original frames for evaluation.

To mitigate overfitting in the Transformer encoder, we applied a dropout rate of 0.2 to the positional embeddings, the attention weights, and the output of each block. 
We also set the drop path rate for the residual connections to 0.2.

The model was trained using the AdamW optimizer with a batch size of 4. 
The optimizer hyperparameters were set to $\beta_1$=0.9, $\beta_2$=0.999, $\epsilon=1\times10^{-8}$, and weight decay  $\lambda=1\times10^{-2}$. 
We employed CosineLRScheduler, a cosine learning rate scheduler with a linear warmup, setting the total number of epochs to 50, the minimum learning rate to 0, the warmup duration to 3 epochs, and the initial warmup learning rate to $1.0\times10^{-5}$.

For the loss function, we utilized the focal loss~\cite{lin2017focal} with the class weights set to $\alpha_{\mathrm{glass}} = 1.56$, $\alpha_{\mathrm{ghost}} = 1.13$, $\alpha_{\mathrm{object}} = 0.536$, and $\alpha_{\mathrm{noise}} = 0.0625$, and the focusing parameter $\gamma$ set to $2.0$.

All training and processing were performed on a computer equipped with an Intel Core i9-14900K CPU and a single RTX 6000 Ada Generation GPU, running the Ubuntu 20.04 operating system.
In this environment, the training time was approximately 5.5 minutes per epoch.

\subsection{Inference}
Because the model input is limited to one-quarter of the full frame, processing an entire frame in a single forward pass is not feasible.
To evaluate the full-frame performance, we followed Ikeda et al.~\cite{ikeda2026ghostfwl} by sequentially applying a non-overlapping sliding window to infer local regions. 
The resulting outputs were then spatially concatenated to reconstruct the full-frame predictions. 
Furthermore, following Ikeda et al.~\cite{ikeda2026ghostfwl} for label assignment, a voxel was classified into the category with a predicted probability exceeding $0.5$; if no class probability exceeded this threshold, it was labeled as \textit{Undefined}.

\subsection{Baseline Methods}
The baseline methods evaluated in our experiments are primarily based on their official implementations, with appropriate modifications to accommodate our specific task and the 3D data structure.
The specific adaptations and settings for each method are detailed below.

\noindent\textbf{Scheuble et al.~\cite{scheuble2025lidar}}: 
As the original model targets a different task, we appended a new prediction head for voxel-wise classification. 
After upsampling the decoder output features to the original spatio-temporal resolution, we applied $1 \times 1\times 1$ 3D convolution to generate the class logits for each voxel.

\noindent\textbf{Ikeda et al.~\cite{ikeda2026ghostfwl}}: 
The architecture and training procedure strictly follow the official implementation. 
However, because this model divides the spatial dimensions into $16 \times 16$ patches, the input spatial resolution must be divisible by 16. 
Therefore, we adjusted the input size from $(H \times W \times T) = (168 \times 200 \times 300)$ used in our method to $(H \times W \times T) = (176 \times 208 \times 300)$ for evaluation.

\noindent\textbf{DynamicViT~\cite{rao2021dynamicvit}}:
We extended this 2D image-based method to accommodate 3D FWL data. 
To align the conditions with our proposed pruning strategy, we configured the model to perform token selection at the input stage and set the keeping ratio to 0.3. 
This ensures that 70\% of the input tokens are pruned, matching our pruning ratio.

\noindent\textbf{ToMe~\cite{bolya2023token}}: 
Similar to DynamicViT, we extended this method for 3D data. 
The original algorithm alternately assigns tokens as dst tokens and src tokens on a 2D grid; we adapted this mechanism to alternate across a 3D token grid.
This structural constraint limits the maximum merge ratio to 0.5.
Because our proposed method utilizes a merge ratio of 0.9, we set the merge ratio in ToMe to its maximum allowable value of 0.5 to approximate our experimental conditions. 
To align with our setup, we applied the merging process to the attention layers. 
Additionally, mirroring our merging strategy, we incorporated a step to copy the features of the dst tokens back to the merged src tokens after the attention computation to maintain consistency with the residual connections.

\section{FWL Generalization Benchmark}

\begin{figure}[tp]
  \begin{center}
    \includegraphics[width=0.9\linewidth]{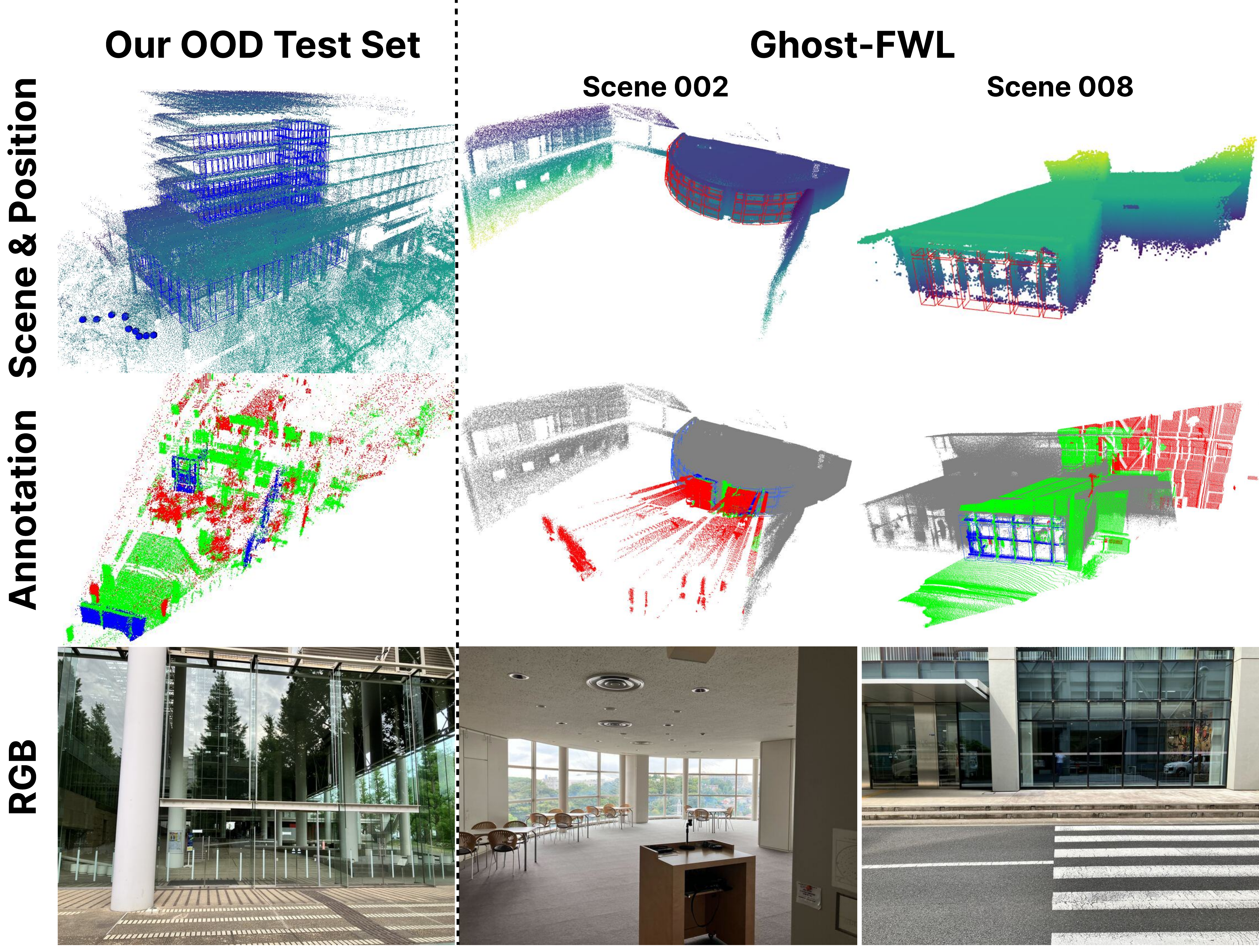}
    \caption{Our OOD Test Set. Based on the dense 3D maps as shown in Scene, we annotated FWL data with 4 labels: Ghost (red), Object (green), Glass (blue), Noise. Blue crosses indicate the LiDAR positions during data acquisition. The RGB images show the scenery of the capture locations.}
    \label{fig:test_set}
  \end{center}
\end{figure}

To evaluate generalization to unseen scenes acquired with the same sensor platform, we construct a new FWL test set distinct from Ghost-FWL~\cite{ikeda2026ghostfwl}. Data were collected using the same FWL sensor as Ghost-FWL, ensuring that any performance difference reflects scene distribution rather than sensor variation. The set consists of 500 frames collected from 10 viewpoints (50 frames each), annotated following the procedure of Ikeda et al.~\cite{ikeda2026ghostfwl}. All scenes were captured outdoors during daytime at a seven-story building clad in glass on all four sides, with additional glass installations around elevators and connecting corridors, totaling over 300 glass windows. Such a glass-dense environment does not appear in Ghost-FWL~\cite{ikeda2026ghostfwl}, making this test set a challenging out-of-distribution benchmark (Figure~\ref{fig:test_set}).

\section{Additional Experiments and Results}

\subsection{Statistical Significance of FWL-Based Comparison}
\label{ssec:statistical_significance_fwl}

Table~\ref{tab:fwl_based_comparison} presents the comparison with existing FWL-based methods 
under different random seeds.
To avoid drawing conclusions from a single trial, we evaluated each method over 3 independent runs 
and report the mean and standard deviation of F1-mean.
This evaluation captures seed-dependent variability and demonstrates that our method achieves superior performance robustly, rather than only under a particular random seed.

The results on the OOD test set further suggest the applicability of FWLAug to unseen data. While Ours fails to detect glass regions located behind other glass surfaces, Ours + FWLAug partially detects such regions, albeit with some inaccuracies. This indicates that FWLAug improves robustness to configurations not observed in the training data.

\begin{table}[t]
\centering
\caption{
Statistical significance of the main comparison with FWL-based methods.
F1-mean is reported as mean~$\pm$~standard deviation over 3 independent runs with different random seeds.
All methods are trained without FWLAug.
}
\label{tab:fwl_based_comparison}
\begin{tabular}{lccccc}
\hline
Method & F1-mean & Object-F1 & Glass-F1 & Ghost-F1 & FPS \\
\hline
Scheuble et al.~\cite{scheuble2025lidar} & 0.497 $\pm$ 0.043 & 0.648 & 0.304 & 0.541 & 1.97 \\
Ikeda et al.~\cite{ikeda2026ghostfwl} & 0.503 $\pm$ 0.005 & 0.680 & 0.268 & 0.561 & 6.54 \\
Ours & \textbf{0.592 $\pm$ 0.005} & \textbf{0.741} & \textbf{0.358} & \textbf{0.678} & \textbf{53.2} \\
\hline
\end{tabular}
\end{table}

\subsection{Comparison with Top-$K$ Peak Extraction}
\begin{figure}[t]
  \centering
  \includegraphics[width=0.9\linewidth]{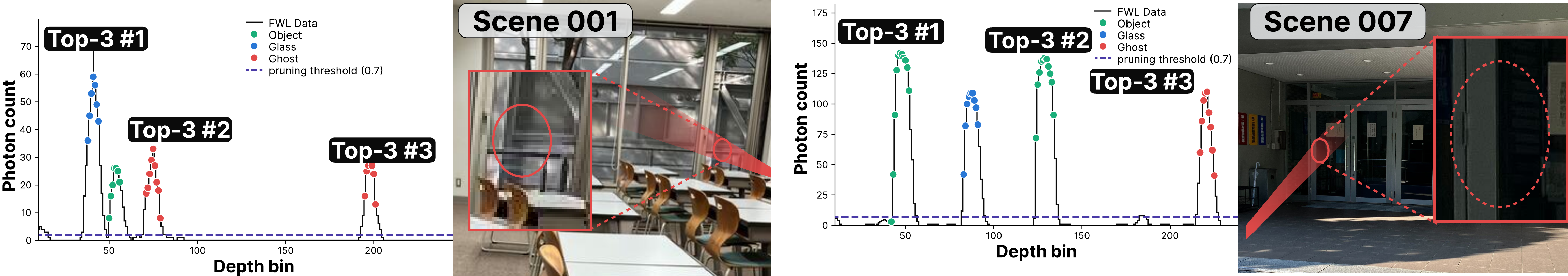}
  \caption{Waveforms along single rays in Ghost-FWL Scene~001 (left) and Scene~007 (right). Markers indicate annotated Object (green), Glass (blue), and Ghost (red) bins, labels indicate the Top-3 highest-intensity peaks of each ray, and the dashed line indicates the intensity threshold below which patches are pruned at $\rho_{\mathrm{low}}=0.7$. The photographs show the corresponding scenes, with the red marker indicating the observed ray. Top-3 peak extraction discards the Object and the Glass return, whereas our intensity-based pruning retains both because they lie above the pruning threshold.}
  \label{fig:waveforms_discard_important_peaks}
\end{figure}

Conventional multi-return ToF LiDAR keeps only the Top-$K$ highest-intensity peaks of each ray. In contrast, FWL-ToPM ranks 3D patches by their maximum intensity over the entire frame, removes a fixed proportion ($\rho_{\mathrm{low}}=0.7$) of low-ranked patches, and retains the complete temporal waveform within each remaining patch. Figure~\ref{fig:waveforms_discard_important_peaks} shows waveforms along single rays from two Ghost-FWL scenes. In Scene~001, the Top-3 peaks consist of a Glass return and two Ghost returns, so the Object return around depth bin 60 is discarded. In Scene~007, two Object returns and a Ghost return occupy the Top-3 peaks, so the Glass return around depth bin 90 is discarded. Since the Top-2 peaks are a subset of the Top-3 peaks, these returns are also discarded by Top-2 extraction. In both cases, the discarded returns lie well above the pruning threshold and are therefore retained by our pruning. These examples illustrate that a fixed per-ray peak budget can drop important Object and Glass returns when strong Ghost returns are present along the same ray, whereas our patch-level pruning preserves them together with their full waveforms.

\subsection{Ablation on Space-Filling Curves}
In FWL-ToPM, the tokens remaining after pruning are serialized along a space-filling curve before dst token selection, and our experiments adopt the Morton order as this curve. To examine whether the choice of curve affects performance, we replace the Morton order with the Hilbert curve~\cite{hilbert} while keeping all other settings unchanged, and evaluate both the merging-only and the pruning + merging settings. As shown in Table~\ref{tab:sfc_ablation}, the two curves yield nearly identical F1-mean scores in both settings, with the Morton order slightly higher: 0.565 versus 0.564 for merging only, and 0.592 versus 0.589 for pruning + merging. This indicates that the benefit comes from serializing the sparse tokens in an order that preserves spatial proximity, rather than from a particular choice of curve.

\begin{table}[t]
\centering
\caption{Comparison of space-filling curves used to serialize tokens for merging. F1 scores on the Ghost-FWL test set are reported. We set $\rho_{\mathrm{low}}=0.7$ (pruning + merging only), $\rho_m=0.9$, and $\rho_d=0.1$. All models are trained without FWLAug.}
\label{tab:sfc_ablation}
\begin{tabular}{llcccc}
\hline
Setting & Curve & Object-F1 & Glass-F1 & Ghost-F1 & F1-mean \\
\hline
\multirow{2}{*}{Merging} & Hilbert & 0.690 & 0.349 & \textbf{0.652} & 0.564 \\
 & Morton (Ours) & \textbf{0.702} & \textbf{0.369} & 0.624 & \textbf{0.565} \\
\hline
\multirow{2}{*}{Pruning + Merging} & Hilbert & 0.740 & 0.355 & 0.672 & 0.589 \\
 & Morton (Ours) & \textbf{0.741} & \textbf{0.358} & \textbf{0.678} & \textbf{0.592} \\
\hline
\end{tabular}
\end{table}

\subsection{Qualitative Evaluation}
\label{ssec:qualitative}
\begin{figure}[tp]
  \begin{center}
    \includegraphics[width=0.8\linewidth]{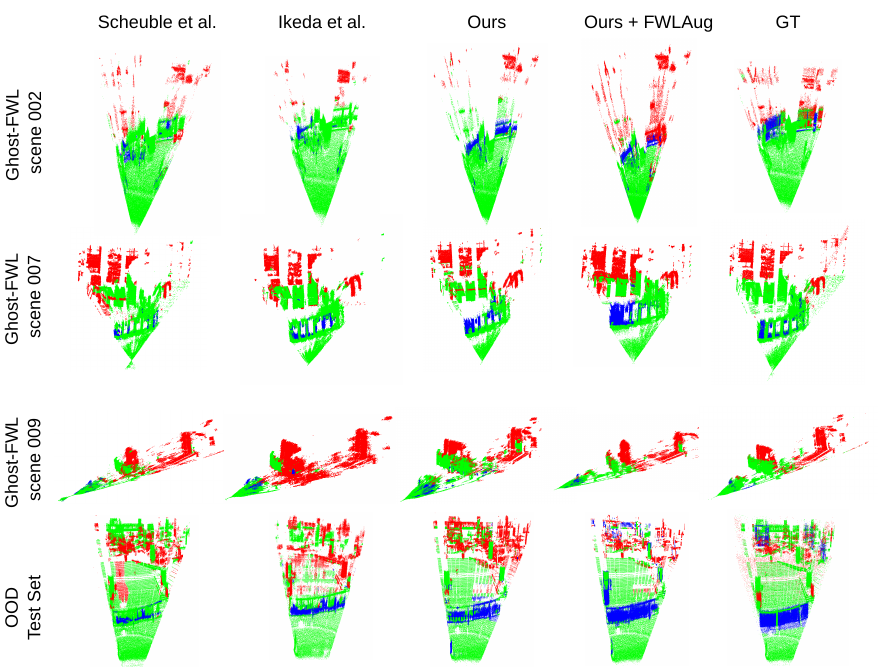}
    \caption{Qualitative comparison of classification results with existing FWL-based recognition methods.}
    \label{fig:qualitative}
  \end{center}
\end{figure}

Figure~\ref{fig:qualitative} shows a qualitative comparison of the classification results with existing FWL-based recognition methods.
We evaluate the methods on scenes from the Ghost-FWL test split and on additional scenes not included in Ghost-FWL~\cite{ikeda2026ghostfwl}.
Across these scenes, Ours classifies Glass and Ghost regions more accurately than the existing methods.
On the additional test scene, Ours also improves the classification of Objects located behind Glass regions, although some Glass regions tend to be partially confused with Object.

Nevertheless, failure cases remain.
In particular, Glass regions located behind other Glass surfaces are still difficult to classify accurately.
We consider this to be partly due to the limited number of such configurations in Ghost-FWL~\cite{ikeda2026ghostfwl}.
By increasing the occurrence of Glass-containing regions during training, FWLAug provides additional waveform and spatial variations associated with Glass.
As a result, Ours+FWLAug shows partial improvements in detecting Glass regions in such challenging configurations, although the predictions remain imperfect.
This suggests that FWLAug can improve robustness to glass configurations that are under-represented in the original training data.

\clearpage
{
\small
\bibliographystyle{plainnat}
\bibliography{supp}
}